\documentclass[letterpaper, 10 pt, conference]{ieeeconf}  % Comment this line out if you need a4paper

\IEEEoverridecommandlockouts                              % This command is only needed if 
\usepackage{graphicx}
\usepackage{url}  %IVM added this because of the URLs in the reference list

\usepackage{amsmath}

\title{\LARGE \bf
An Audio-Visual Dataset and Deep Learning Frameworks \\ for Crowded Scene Classification
}

\author{Lam~Pham$^{1}$,
        Dat~Ngo$^{2}$,
        Phu~X.~Nguyen$^{3}$,
        Truong~Hoang$^{4}$,
        Alexander~Schindler$^{1}$ %
\thanks{L. Pham and A. Schindler are with Competence Unit Data Science \& Artificial Intelligence, Center for Digital Safety \& Security, Austria Institute of Technology, Austria.}%
\thanks{D. Ngo is with School of Computer Science and Electronic Engineering, University of Essex, UK.}%
\thanks{Phu   X.   Nguyen   is   with   Department   of   Computer   Fundamentals,  FPT  University,  Ho  Chi  Minh  City  700000,  Vietnam.}
\thanks{T. Hoang is with FPT Software Company Limited, Vietnam.}%
}

\begin{document}

\maketitle
\thispagestyle{empty}
\pagestyle{empty}

%%%%%%%%%%%%%%%%%%%%%%%%%%%%%%%%%%%%%%%%%%%%%%%%%%%%%%%%%%%%%%%%%%%%
\begin{abstract}
This paper presents a task of audio-visual scene classification (SC) where input videos are classified into one of five real-life crowded scenes: `Riot', `Noise-Street', `Firework-Event', `Music-Event', and `Sport-Atmosphere' . 
To this end, we firstly collect an audio-visual dataset (videos) of these five crowded contexts from Youtube (in-the-wild scenes).  
Then, a wide range of deep learning frameworks are proposed to deploy either audio or visual input data independently.
Finally, results obtained from high-performed deep learning frameworks are fused to achieve the best accuracy score.   
Our experimental results indicate that audio and visual input factors independently contribute to the SC task's performance.
Significantly, an ensemble of deep learning frameworks exploring either audio or visual input data can achieve the best accuracy of 95.7\%.   
\newline

\indent \textit{Technical terms}--- Deep learning framework, convolutional neural network (CNN), scene classification (SC), data augmentation.
\end{abstract}
%%%%%%%%%%%%%%%%%%%%%%%%%%%%%%%%%%%%%%%%%%%%%%%%%%%%%%%%%%%%%%%%%%%%%%%%%%%%%%%%

\section{INTRODUCTION}
\label{intro}
The work presented in this paper is a part of our project which arms to deal with alarming events such as protests or riots.
In particular, the project leverages Artificial Intelligent (AI) techniques to be able to detect these alarming events early and automatically before these events are reported on main streams (e.g. Radio or TV channels).
By early detecting relevant-riot contexts (e.g. which/where/when the event occur?), we can predict a possible large-scale migration or immediately trigger a warning for a certain region (e.g. A violent riot is occurring at the street/district/country X).
To this end, the updated data (e.g. text, audio, and image) extracted from posts on various social networks (e.g. Twitter, Facebook, Youtube, etc.) are firstly collected.
Then, text, audio, and image input data are automatically analysed by independent AI-based models.
The results obtained from multiple models are finally fused, then consider whether a warning should be triggered.

In this paper, we focus on analysing videos (audio and visual input data), then indicate whether videos are close to a riot context. 
To this end, we define a scene classification (SC) task in this paper where real-life crowded scenes, which includes the riot or protest context, are classified.
Regarding current scene classification tasks, it can be seen that almost public datasets have been proposed for detecting daily scenes rather than specific crowded scenes.
For examples, DCASE Task 1 challenges from 2018 to 2021 datasets~\cite{dcase_web} present 10 daily audio or audio-visual scenes of `bus', `tram', `metro', `public square', `park', `pedestrian street', `traffic street', `tram station', `airport', `shopping mall', or ESC-50~\cite{esc50_ds} dataset presents 5 audio scene categories of `relevant animal', `natural context', `relevant human', `interior/domestic context', and `exterior/urban noise context'. 
Although some audio-visual datasets have been proposed to detect violence such as XD-Violence~\cite{xd_vio_ds} or UCF-Crime~\cite{ucf_dataset} which are close to riot or protest contexts, a violent scene may not occur in certain peaceful protests. 
Even, singing and clapping sounds, which are frequently in some peaceful protests, can make classification models easily misclassify into a context of music events.    
Furthermore, it is vital that there are some contexts such as sport atmosphere (i.e. A sport atmosphere in a stadium presents a very noise and crowded scene) or firework events (i.e. A firework event presents a crowded scene with a lot of cracker and very similar gun sounds), which are very close to a riot context and easy to be misclassified together.  
To deal with the issue of lacking dataset, we firstly collect an audio-visual dataset (videos) from Youtube (in-the-wild scenes), which comprises five crowded scenes of `Riot', `Noise-Street', `Firework-Event', `Music-Event', and `Sport-Atmosphere'.
Then, we leverage deep learning techniques, which show powerful to deal with SC tasks~\cite{xd_vio_ds, aud_vis_ana, lam01, lam02}, to analyse this dataset and then indicate: \textbf{(1)} whether it is possible to achieve a high-performed framework for classifying these five crowded scenes which is very important and potential for the whole project mentioned; and \textbf{(2)} whether each audio and visual input factor independently contributes to the task of classifying crowded scenes defined.

\section{An Audio-Visual dataset of Five Crowded Scenes and Task Definition}
\label{dataset}
%++++++++++++++
\begin{figure*}[ht]
    \centering
    \includegraphics[width =1.0\linewidth]{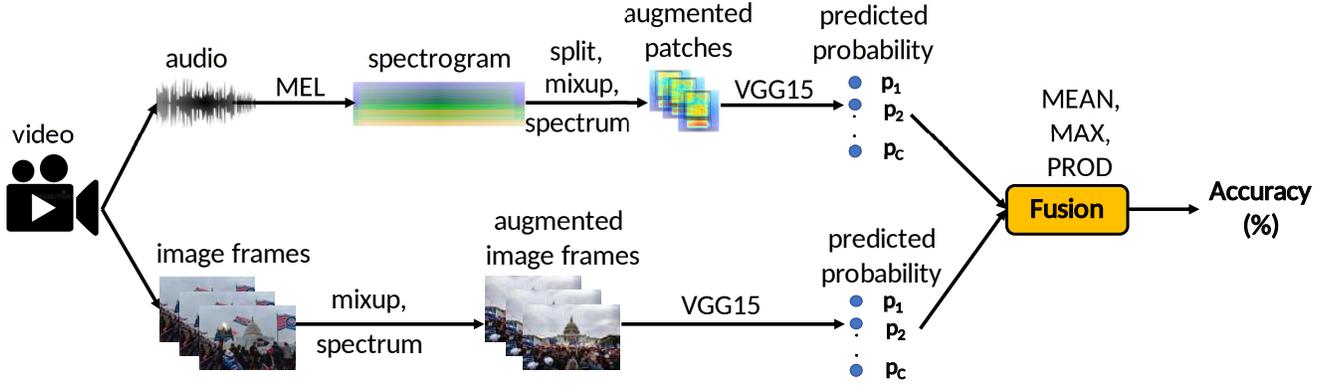}
    	\vspace{-0.5cm}
	\caption{The high-level architecture of the aud-vis baseline proposed for classifying five crowded scenes.}
    \label{fig:baseline}
\end{figure*}
%++++++++++++++
Our dataset of 341 videos were collected from YouTube (in-the-wild scenes), which presents a total recording time of nearly 29.06 hours.
These videos are then split into 10-second video segments, each which is annotated by one of five categories of `riot', `noise street', `firework event',`music event', or `sport atmosphere'.
The video segments are split into two subsets of Train and Test with the ratio of 67:33 for the training and inference processes, respectively\footnote{https://zenodo.org/record/5774751\#.Ybc9R5pKhhE}.
Notably, 10-second video segments split from an original video are not presented in both Train and Test subsets to make the data distribution different on these two subsets. 
The number of 10-second video segments on each subset is shown in Table~\ref{table:dataset}.

%--------------------------------------------
\begin{table}[t]
	\caption{The number of 10-second video segments corresponding to each scene categories in Train. and Test. subsets} 
	%\vspace{-0.2cm}
	\centering
    \scalebox{0.95}{
	\begin{tabular}{|l |c |c |c |c|} 
		\hline 
		\textbf{Category}         & \textbf{Train}     & \textbf{Test}   & \textbf{Total}     \\ 
		\hline 
		Riot 	                & 1429                  & 757   & 2186     \\        
		Noise-Street     	    & 1430                  & 652   & 2082     \\        
		Firework-Event	 	    & 1406                  & 615   & 2021     \\        
		Music-Event         	& 1367                  & 727   & 2094     \\        
		Sport-Atmosphere        & 1365                  & 712   & 2077     \\     
     	\hline 
  		Total                   & 6997                  & 3463  & 10460   \\
                                & ($\approx$ 19.44 hours) & ($\approx$9.62 hours) & ($\approx$29.06 hours) \\
		\hline 
	\end{tabular}  
	}  
	\label{table:dataset} 
   %\vspace{-0.5cm}
\end{table} 
%--------------------------------------------
Regarding the evaluation metric, the Accuracy (Acc.\%), which is commonly applied in SC challenges such as DCASE Task 1A~\cite{dcase_web}, is used to evaluate the task of crowded scene classification in this paper. 
Let us consider $M$ as the number of 10-second video segments, which are correctly predicted, from the total number of segments as $N$, the classification accuracy (Acc.\%) is computed by
\begin{equation}
    \label{eq:c03_mean_stratergy_patch}
    Acc.\%= 100\frac{M}{N}.
\end{equation}
\section{Proposed Deep Learning Frameworks}
\label{frameworks}

\subsection{The audio, visual, and aud-vis baselines proposed}
\label{baseline}

As we aim to analyse the independent impact of audio or visual input data on the classification task's performance, proposed deep learning frameworks deploy either audio or visual input data.
Then, the results obtained from individual high-performed frameworks are fused to achieve the best performance.
To this end, we firstly propose a deep learning framework as described in Figure~\ref{fig:baseline}, referred to as the aud-vis baseline.
As the Figure~\ref{fig:baseline} shows, the audio and visual input data extracted from a video are deployed by two independent streams, referred to as the audio baseline (e.g. the upper data stream) and the visual baseline (e.g. the lower data stream), before a fusion of predicted probability results of audio and visual streams is conducted.

Regarding the audio baseline as shown in the upper stream in Figure~\ref{fig:baseline}, input audio recordings are firstly resampled to 32,000 Hz, then transformed into MEL spectrograms where both temporal and frequency features are presented. 
By using only one channel and setting parameters of filter number, window size, hop size to 128, 80 ms, 14 ms respectively, we generate MEL spectrograms of $640{\times}128$ from 10-second audio segments.
Next, entire spectrograms are split into small patches of $128{\times}128$ that are suitable for back-end classification models.
To enforce back-end classifiers, two data augmentation methods of spectrum~\cite{spec_aug} and mixup~\cite{mixup2} are applied on these patches before feeding into a back-end VGGish network for classification.
Regarding spectrum augmentation, we apply two zero masking blocks of 10 frequency channels and 10 time frames on each patch of  $128{\times}128$.
The starting frequency channel or time frame in a masking block is randomly selected.
The patches of $128{\times}128$ after spectrum augmentation are then mixed together with different ratios, which is known as the mixup data augmentation.
Let's consider two patches of $128{\times}128$ as \(\mathbf{X_{A}}\), \(\mathbf{X_{B}}\) and expected labels as \(\mathbf{y_{A}}\), \(\mathbf{y_{B}}\), we generate new patches as below equations:
\begin{equation}
    \label{eq:mix_up_x1}
    \mathbf{X_{mx1}} = \gamma\mathbf{X_{A}} + (1-\gamma)\mathbf{X_{B}}
\end{equation}
\begin{equation}
    \label{eq:mix_up_x2}
    \mathbf{X_{mx2}} = (1-\gamma)\mathbf{X_{A}} + \gamma\mathbf{X_{B}}
\end{equation}
\begin{equation}
    \label{eq:mix_up_y}
    \mathbf{y_{mx1}} = \gamma\mathbf{y_{A}} + (1-\gamma)\mathbf{y_{B}}
\end{equation}
\begin{equation}
    \label{eq:mix_up_y}
    \mathbf{y_{mx2}} = (1-\gamma)\mathbf{y_{A}} + \gamma\mathbf{y_{B}}
\end{equation}
with \(\gamma\) is random coefficient from unit distribution or gamma distribution.
We feed patches of $128{\times}128$ before and after mixup data augmentation into a back-end VGGish network architecture for classification.    

For the visual baseline as shown in the lower stream in Figure~\ref{fig:baseline}, two data augmentation methods of spectrum~\cite{spec_aug} and mixup~\cite{mixup2} are also applied on visual data (image frames) before feeding into a back-end VGGish network architecture.

The back-end VGGish networks for audio and visual streams are independent, but share the same architecture as described in Table~\ref{table:VGG}.
As Table~\ref{table:VGG} shows, the VGGish network contains sub-blocks which perform convolution (Conv[kernel size]@channel), batch normalization (BN)~\cite{batchnorm}, rectified linear units (ReLU)~\cite{relu}, average pooling (AP), global average pooling (GAP), dropout (Dr(percentage))~\cite{dropout}, fully connected (FC) and Softmax layers. 
The dimension of Softmax layer is set to $C=5$ which corresponds to the number of crowded scenes classified.
In total, we have 12 convolutional layers and 3 fully connected layers containing trainable parameters that makes the proposed network architecture like VGG15~\cite{vgg_net}.
%--------------------------------------------
\begin{table}[t]
    \caption{The VGG15 network architecture used for both audio and visual baselines.} 
        	\vspace{-0.2cm}
    \centering
    \scalebox{0.95}{
    \begin{tabular}{|l |c |} 
        \hline 
            \textbf{Network Architecture}   &  \textbf{Output}  \\
                                            %&  \textbf{(FxTxC)}  \\
        \hline 
         BN - Conv [$3{\times}3] $@$  32$ - ReLU - BN - Dr (20\%) & $128{\times}128{\times}32$\\
         BN - Conv [$3{\times}3] $@$  32$ - ReLU - BN - AP - Dr (25\%) & $64{\times}64{\times}32$\\
         
         BN - Conv [$3{\times}3] $@$ 64$ - ReLU - BN - Dr (25\%) & $64{\times}64{\times}64$\\
         BN - Conv [$3{\times}3] $@$ 64$ - ReLU - BN - AP - Dr (30\%) & $32{\times}32{\times}64$\\
         
         BN - Conv [$3{\times}3] $@$ 128$ - ReLU - BN - Dr (30\%) & $32{\times}32{\times}128$\\
         BN - Conv [$3{\times}3] $@$ 128$ - ReLU - BN - Dr (30\%) & $32{\times}32{\times}128$\\
         BN - Conv [$3{\times}3] $@$ 128$ - ReLU - BN - Dr (30\%) & $32{\times}32{\times}128$\\
         BN - Conv [$3{\times}3] $@$ 128$ - ReLU - BN - AP - Dr (30\%) & $16{\times}16{\times}128$\\
         
         BN - Conv [$3{\times}3] $@$ 256$ - ReLU - BN - Dr (35\%) & $16{\times}16{\times}256$\\
         BN - Conv [$3{\times}3] $@$ 256$ - ReLU - BN - Dr (35\%) & $16{\times}16{\times}256$\\
         BN - Conv [$3{\times}3] $@$ 256$ - ReLU - BN - Dr (35\%) & $16{\times}16{\times}256$\\
         BN - Conv [$3{\times}3] $@$ 256$ - ReLU - BN - GAP - Dr (35\%) & $256$\\
                  
         FC - ReLU - Dr (40\%)&  $1024$       \\
         FC - ReLU - Dr (40\%)&  $1024$       \\
         FC - Softmax  &  $C=5$       \\
       \hline 
           %	\vspace{-1cm}
    \end{tabular}
    }
    \label{table:VGG} 
\end{table}
%--------------------------------------------

As back-end classifiers work on patches of $128{\times}128$ or image frames, the predicted probability of an entire spectrogram or all image frames from a 10-second video segment is computed by averaging of all images or patches' predicted probabilities.
Let us consider $\mathbf{P^{n}} = (\mathbf{p_{1}^{n}, p_{2}^{n},..., p_{C}^{n}})$,  with $C$ being the category number and the \(n^{th}\) out of \(N\) image frames or patches of $128{\times}128$ fed into a deep learning model, as the probability of a test video, then the average classification probability is denoted as  \(\mathbf{\bar{p}} = (\bar{p}_{1}, \bar{p}_{2}, ..., \bar{p}_{C})\) where,
\begin{equation}
    \label{eq:mean_stratergy_patch}
    \bar{p}_{c} = \frac{1}{N}\sum_{n=1}^{N}p_{c}^{n}  ~~~  for  ~~ 1 \leq n \leq N 
\end{equation}
and the predicted label \(\hat{y}\) for an entire spectrogram or all image frames is determined as:

\begin{equation}
    \label{eq:label_determine}
    \hat{y} = arg max (\bar{p}_{1}, \bar{p}_{2}, ...,\bar{p}_{C} )
\end{equation}

To evaluate the aud-vis baseline, an ensemble of results from the individual audio and visual baselines is conducted.
In particular, we proposed three late fusion schemes, namely MEAN, PROD, and MAX fusions.
For each scheme, we firstly conduct experiments on the individual audio and visual baselines, then obtain the predicted probability of each baseline as  \(\mathbf{\bar{p_{s}}}= (\bar{p}_{s1}, \bar{p}_{s2}, ..., \bar{p}_{sC})\) where $C$ is the category number and the \(s^{th}\) out of \(S\) individual frameworks evaluated. 
Next, the predicted probability after MEAN fusion \(\mathbf{p_{f-mean}} = (\bar{p}_{1}, \bar{p}_{2}, ..., \bar{p}_{C}) \) is obtained by:
\begin{equation}
    \label{eq:mix_up_x1}
     \bar{p_{c}} = \frac{1}{S} \sum_{s=1}^{S} \bar{p}_{sc} ~~~  for  ~~ 1 \leq s \leq S 
\end{equation}

The PROD strategy \(\mathbf{p_{f-prod}} = (\bar{p}_{1}, \bar{p}_{2}, ..., \bar{p}_{C}) \) is obtained by:
\begin{equation}
\label{eq:mix_up_x1}
\bar{p_{c}} = \frac{1}{S} \prod_{s=1}^{S} \bar{p}_{sc} ~~~  for  ~~ 1 \leq s \leq S 
\end{equation}
and the MAX strategy \(\mathbf{p_{f-max}} = (\bar{p}_{1}, \bar{p}_{2}, ..., \bar{p}_{C}) \) is obtained by:
\begin{equation}
\label{eq:mix_up_x1}
\bar{p_{c}} = max(\bar{p}_{1c}, \bar{p}_{2c}, ..., \bar{p}_{Sc}) 
\end{equation}
Finally, the predicted label  \(\hat{y}\) is determined by (\ref{eq:label_determine}). 

\subsection{Further exploring audio-based frameworks}
\label{audio_frame}

As applying an ensemble of either different types of input spectrograms~\cite{lam03, lam04, lam05, lam_11} or different learning models~\cite{seo_2019, huang_2019, haocong2019acoustic, truc_dca_18, huy_mul} has been a rule of thumb to enhance the performance of audio-based scene classification task performance, we therefore evaluate two ensemble methods, referred to as the multiple spectrogram strategy (e.g. Multiple spectrograms combines with one model) and the multiple model strategy (e.g. Multiple models deploys one type of spectrogram).
The multiple spectrogram approach uses three types of spectrograms: Constant Q Transform (CQT)~\cite{librosa_tool}, Mel filter (MEL)~\cite{librosa_tool}, and Gammatone filter (GAM)~\cite{aud_tool}. 
Each spectrogram is then independently classified by one VGG15 as described in Table~\ref{table:VGG}.
We refer to three frameworks as \textit{CQT-VGG15, GAM-VGG15}, and \textit{MEL-VGG15} (i.e. MEL-VGG15 is known as the audio baseline mentioned in Section~\ref{baseline}), respectively.
In the multiple model approach, while only MEL spectrogram is used, different back-end classifiers are evaluated.
In particular, we use five benchmarks deep neural network architectures from Keras application library~\cite{keras_app}: Xception, Resnet50, InceptionV3, MobileNet, and DenseNet121. 
We refer to these five frameworks as \textit{MEL-Xception, MEL-Resnet50, MEL-InceptionV3, MEL-MobileNet}, and  \textit{MEL-DenseNet121}, respectively.
In these both approaches, the final classification accuracy is obtained by applying late fusion methods (MAX, MEAN, and PROD) of individual frameworks as mentioned in Section~\ref{baseline} (i.e. An ensemble of three predicted probabilities from \textit{CQT-VGG15, GAM-VGG15}, and \textit{MEL-VGG15}, or an ensemble of five predicted probabilities from \textit{MEL-Xception, MEL-Resnet50, MEL-InceptionV3, MEL-MobileNet}, and \textit{MEL-DenseNet121}).

\subsection{Further exploring visual-based frameworks}
\label{visual_frame}
%++++++++++++++
\begin{figure*}[t]
    \centering
    \includegraphics[width =1.0\linewidth]{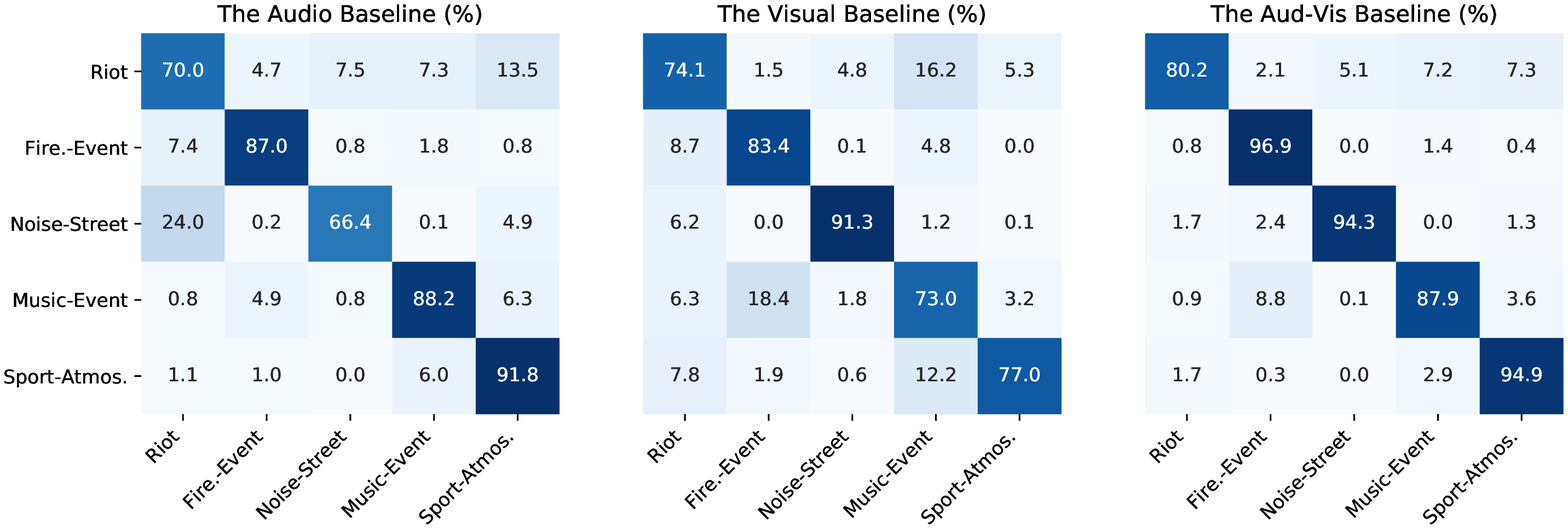}
    	\vspace{-0.5cm}
	\caption{Confusion matrix results of the audio baseline, visual baseline, and aud-vis baseline using PROD fusion.}
    \label{fig:F1}
\end{figure*}
%++++++++++++++

To further exploring visual-based frameworks, we replace VGG15 by five different network architectures from Keras application library~\cite{keras_app}: Xception, Resnet50, InceptionV3, MobileNet, and DenseNet121 which are same and mentioned in Section~\ref{audio_frame}.
Notably, the output nodes at the final fully connected layer of these network architectures is modify from 1000 (e.g. The number of classes in Imagenet dataset) to 5 which matches the number of crowded scene categories. 
Additionally, we propose two training strategies: Direct Training and Fine-tuning, which are applied on these five network architectures.
In Direct Training strategy, all trainable parameters of these five network architectures are initialized with zero and variance set to 0 and 0.1 respectively.
Meanwhile in the Fine-tuning strategy, these five networks were trained with ImageNet dataset~\cite{Imagenet} in advance.
We then only reuse the trainable parameters from the first layer to the global pooling layer and train the entire network with a low learning rate of 0.00001.
The visual-based frameworks with Direct Training strategy are referred to as \textit{visual-direct-VGG15} (i.e. \textit{Visual-direct-VGG15} is known as the visual baseline) and \textit{visual-direct-Xception, visual-direct-Resnet50, visual-direct-InceptionV3, visual-direct-MobileNet}, \textit{visual-direct-DenseNet121}, respectively.
Meanwhile, the visual-based frameworks with Fine-tuning strategy are referred to as \textit{visual-finetune-Xception, visual-finetune-Resnet50, visual-finetune-InceptionV3, visual-finetune-MobileNet}, and \textit{visual-finetune-DenseNet121}, respectively.
Similar to the audio-based frameworks, the final classification accuracy of the visual-based frameworks is obtained by applying late fusion methods (MAX, MEAN, PROD) of individual frameworks (i.e. An ensemble of predicted probabilities from all frameworks with either Direct Training or Fine-tuning strategies).

\subsection{Implementation of deep learning frameworks}

We use Tensorflow framework to build all classification models in this paper.
As we apply spectrum~\cite{spec_aug} and mixup~\cite{mixup2} data augmentation for both audio spectrograms and image frames to enforce back-end classifiers, the labels of the augmented data are no longer one-hot.
We therefore train back-end classifiers with Kullback-Leibler (KL) divergence loss~\cite{kl_loss} rather than the standard cross-entropy loss over all $N$ training samples:
\begin{align}
    \label{eq:loss_func}
    LOSS_{KL}(\Theta) = \sum_{n=1}^{N}\mathbf{y}_{n}\log\left(\frac{\mathbf{y}_{n}}{\mathbf{\hat{y}}_{n}} \right)  +  \frac{\lambda}{2}||\Theta||_{2}^{2},
\end{align}
where $\Theta$ denotes the trainable network parameters and $\lambda$ denotes the $\ell_2$-norm regularization coefficient. $\mathbf{y_{c}}$ and $\mathbf{\hat{y}_{c}}$  denote the ground-truth and the network output, respectively. 
The training is carried out for 100 epochs using Adam method~\cite{Adam} for optimization.
All experiments are running on the GeForce RTX 2080 Titan GPUs.

\section{Experimental Results and Discussions}
\label{result}

\subsection{Performance comparison of the audio baseline, visual baseline, and aud-vis baseline}
\label{baseline_fac}
As the confusion matrix results of the audio baseline, the visual baseline, and the aud-vis baseline with PROD fusion are shown in Figure~\ref{fig:F1}, we can see that the audio baseline and the visual baseline are very competitive on `Riot' and `Firework-Event' classes, but they show significant gaps of performance in `Noise-Street', `Music-Event', and `Sport-Atmosphere' categories.
 
When we apply PROD fusion on predicted probabilities of the audio and visual baselines (e.g. the aud-vis baseline with PROD fusion), performance of all scene categories are significantly improved.
This proves that audio or visual input factors have distinct and independent contribution on the task of crowded scene classification proposed.

\subsection{Performance comparison of audio-based frameworks}
\label{aud_fac}
%++++++++++++++
\begin{figure*}[t]
    \centering
    \includegraphics[width =1.0\linewidth]{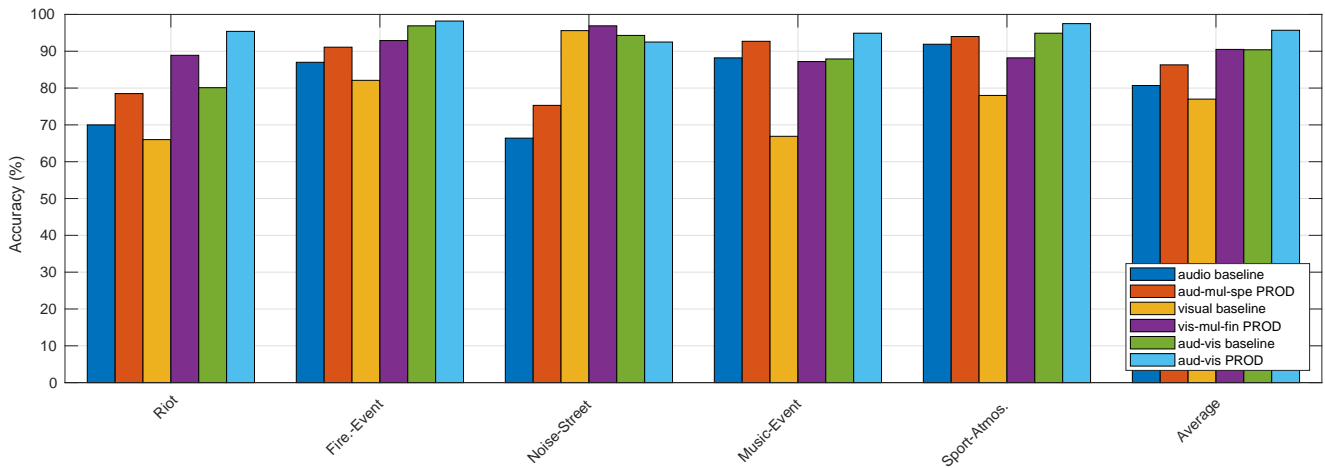}
    	\vspace{-0.5cm}
	\caption{Performance comparison (Acc.\%) of \textbf{audio baseline} (\textit{MEL-VGG15}), \textbf{aud-mul-spe PROD} (PROD fusion of \textit{CQT-VGG15, GAM-VGG15, MEL-VGG15}), \textbf{visual baseline} (\textit{visual-direct-VGG15}), \textbf{vis-mul-fin PROD} (PROD fusion of \textit{visual-finetune-Xception, visual-finetune-Inception3,} and \textit{visual-finetune-DenseNet121}), \textbf{aud-vis baseline} (PROD fusion of \textit{MEL-VGG15} and \textit{visual-direct-VGG15}), and \textbf{aud-vis PROD} (PROD fusion of \textit{CQT-VGG15, GAM-VGG15, MEL-VGG15} and  \textit{visual-finetune-Xception, visual-finetune-InceptionV3, visual-finetune-DenseNet121}) across all scene categories.}
    \label{fig:F5}
        	%\vspace{-1cm}
\end{figure*}
%++++++++++++++
As the performance of audio-based frameworks are shown in Table~\ref{table:au_res}, we can see that the ensembles of multiple spectrogram or multiple model frameworks help to improve the performance, present the best score of 86.3\% and 85.8\% from PROD fusion of \textit{CQT-VGG15, GAM-VGG15, MEL-VGG15} and MAX fusion of \textit{MEL-VGG15, MEL-Xception, MEL-Resnet50, MEL-InceptionV3, MEL-MobileNet, MEL-DenseNet121}.
These results also indicate that although multiple spectrogram approach uses the low footprint network architecture of VGG15, this approach is more effective than the multiple model approach with high-complexity networks.

Analysing performance on each spectrogram, GAM and MEL achieve competitive results of 83.0\% and 80.7\% respectively. 
Meanwhile, CQT shows a slightly poorer performance of 78.6\%.
However, while a PROD fusion of \textit{GAM-VGG15} and \textit{MEL-VGG15} achieves 84.2\%, PROD fusion of all three spectrograms helps to further improve the performance by 2.1\%. This proves that each spectrogram contains distinct features of audio scenes.

For the multiple model approach, it can be seen that individual deep learning frameworks show competitive performance, present the lowest and highest scores of 79.8\% and 83.1\% from \textit{MEL-MobileNet} and \textit{MEL-DenseNet121}, respectively.

\subsection{Performance comparison of visual-based frameworks}
\label{vis_fac}

As the performance of visual-based frameworks is shown in Table~\ref{table:vi_res}, we can see that deep learning frameworks with Fine-tuning strategy significantly outperform the same network architectures applying Direct Training strategy.
With the Fine-tuning strategy, we can achieve the best accuracy of 88.4\% from the single \textit{visual-finetune-Xception} or \textit{visual-finetune-InceptionV3} frameworks. 
Meanwhile, the best performance of individual frameworks with Direct Training strategy is 80.7\% from \textit{visual-finetune-Resnet50}.

Ensembles of deep learning frameworks regarding both training strategies only help to improve the performances slightly.
The results record an improvement of 1.5\% and 1.3\% for Direct Training and Fine-tuning compared with the best single frameworks of \textit{visual-direct-Resnet50} and \textit{visual-finetune-Xception}, respectively. 

\subsection{Performance comparison among audio-based, visual-based, and audio-visual frameworks}
\label{all_fac}

%++++++++++++++
\begin{table}[t]
	\caption{Performance of audio-based frameworks.} 
	\vspace{-0.2cm}
	\centering
    \scalebox{0.95}{
	\begin{tabular}{|l |c |l |c |} 
		\hline 
		\textbf{Multiple Spectrogram}  & \textbf{Acc.\%}     & \textbf{Multiple Model}    & \textbf{Acc.\%}   \\ 
		\textbf{Approach}        &                       & \textbf{Approach}  &                 \\ 
		\hline 
		MEL-VGG15 (audio baseline) & 80.7                   & MEL-MobileNet       &    79.8 \\        
		GAM-VGG15     	           & \textbf{83.0}          & MEL-Resnet50        &    81.4 \\        
		CQT-VGG15                  & 78.6                   & MEL-Xception        &    81.6 \\        
		                           &                        & MEL-InceptionV3     &    81.7 \\   
		                           &                        & MEL-DenseNet121     &    83.1 \\   
  		\hline 
		MAX Fusion          & 85.5                   & MAX Fusion    &  \textbf{85.8}  \\        
		MEAN Fusion 	    & 85.8                   & MEAN Fusion   &  85.0  \\        
		PROD Fusion         & \textbf{86.3}          & PROD Fusion   &  85.5 \\        
		\hline 
	\end{tabular}    
	}
    \vspace{-0.2cm}
	\label{table:au_res} 
\end{table}
%++++++++++++++

As the comprehensive analysis of independent audio and visual input factors shows in Section~\ref{aud_fac} and~\ref{vis_fac}, it can be seen that multiple spectrogram approach for audio input data and Fine-tuning strategy for visual input data are effective to enhance the the SC task's performance.
We now combine both visual and audio input factor, then compare: \textbf{(I)} \textit{MEL-VGG15} framework known as the \textbf{audio baseline}, \textbf{(II)} PROD fusion of \textit{CQT-VGG15, GAM-VGG15, MEL-VGG15} referred to as the \textbf{aud-mul-spe PROD}, \textbf{(III)} \textit{visual-direct-VGG15} framework known as the \textbf{visual baseline}, \textbf{(IV)} PROD fusion of \textit{visual-finetune-Xception, visual-finetune-InceptionV3} and \textit{visual-finetune-DenseNet121} referred to as the \textbf{vis-mul-fin PROD}, \textbf{(V)} PROD fusion of \textit{MEL-VGG15} and \textit{visual-direct-VGG15} known as the \textbf{aud-vis baseline}, and \textbf{(VI)} PROD fusion of \textit{CQT-VGG15, GAM-VGG15, MEL-VGG15} and  \textit{visual-finetune-Xception, visual-finetune-InceptionV3, visual-finetune-DenseNet121} referred to as the \textbf{aud-vis PROD} (i.e. \textbf{(VI)} is PROD fusion of \textbf{(II)} and \textbf{(IV)}) across all crowded scene categories.

As the results shows in Figure~\ref{fig:F5}, it can be seen that while the audio baseline (\textit{MEL-VGG15}) and the visual baseline (\textit{visual-direct-VGG15}) show very competitive average scores of 80.7\% and 79.3\% respectively, the ensemble of the best three visual based frameworks (vis-mul-fin PROD with 90.5\%) outperforms the ensemble of multiple-spectrogram audio based frameworks (aud-mul-spe PROD with 86.3\%).
%++++++++++++++
\begin{table}[t]
    \caption{Performance of visual-based frameworks.} 
        	\vspace{-0.2cm}
    \centering
    \scalebox{0.85}{
    \begin{tabular}{|l |c |c |} 
        \hline 
        \textbf{Deep Learning Frameworks}         & \textbf{Direct Training }    & \textbf{Fine-tuning}   \\ 
		                                          & \textbf{Strategy (Acc.\%)}            & \textbf{Strategy (Acc.\%)}  \\ 		
        \hline 
         visual-direct-VGG15 (visual baseline)& 79.3  & - \\
         visual-direct/finetune-MobileNet     & 78.3  & 86.9 \\
         visual-direct/finetune-Resnet50      & \textbf{80.7}  & 84.8 \\
         visual-direct/finetune-Xception      & 79.4  & \textbf{88.4} \\
         visual-direct/finetune-InceptionV3   & 77.8  & \textbf{88.4} \\
         visual-direct/finetune-DenseNet121   & 79.6  & 87.0 \\         
        \hline 
        MAX Fusion     & 81.6   & 89.3      \\        
		MEAN Fusion    & 82.1   & 89.5      \\        
		PROD Fusion    & \textbf{82.2}   & \textbf{89.7}      \\    		    
		        \hline 
           %\vspace{-0.2cm}
    \end{tabular}
    }
    \label{table:vi_res} 
\end{table}
%++++++++++++++
%++++++++++++++
\begin{figure*}[t]
    \centering
    \includegraphics[width =0.92\linewidth]{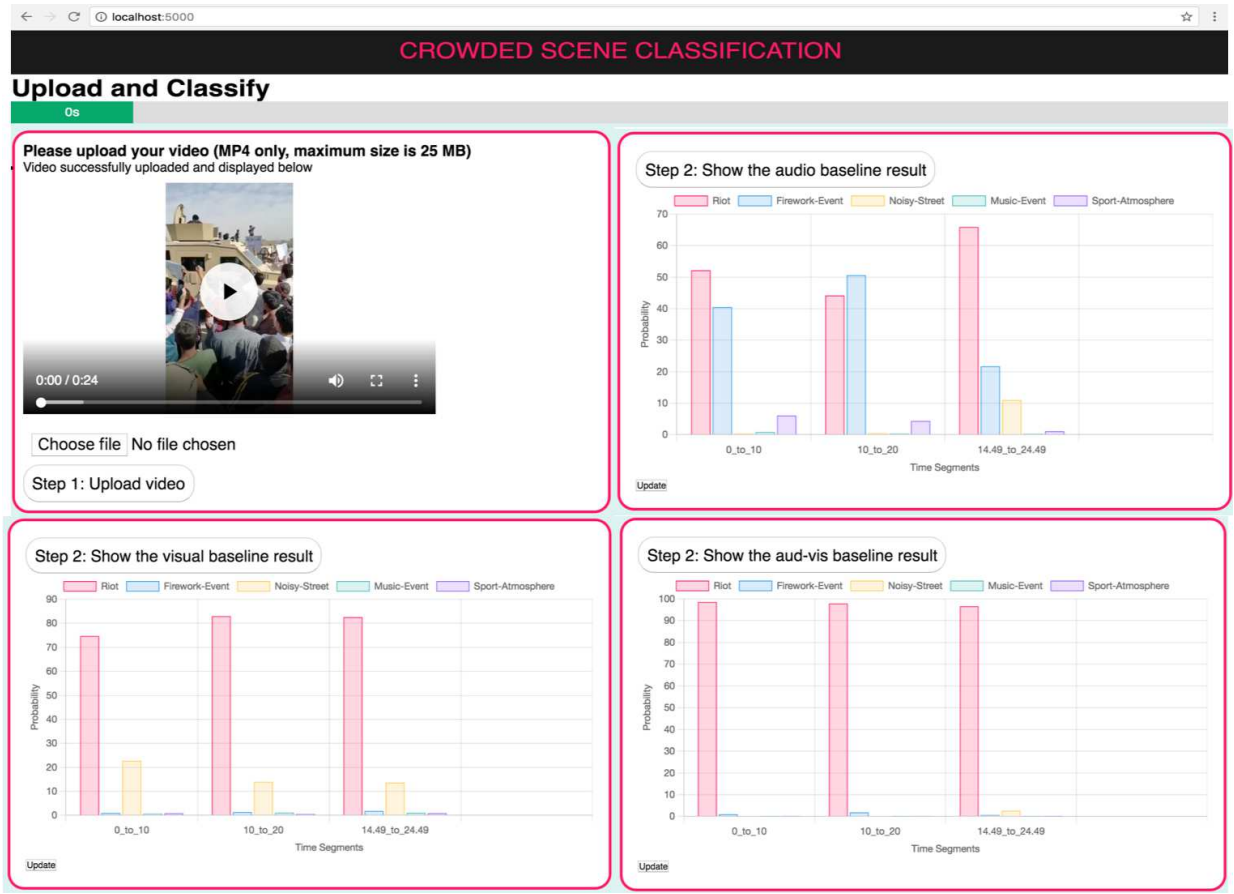}
    	\vspace{-0.2cm}
	\caption{An application demo for audio-visual crowded scene classification.}
    \label{fig:F6}
\end{figure*}
%++++++++++++++

Compare performance between the aud-vis baseline (e.g. PROD fusion of \textit{MEL-VGG15} trained on audio input and \textit{visual-direct-VGG15} trained on visual input) and vis-mul-fin PROD (i.e. This approach makes use of Fine-tuning strategy with three large networks of Xception, InceptionV3 and DenseNet121 which are trained with only image input data), it proves that an ensemble of both audio and visual data with the VGG15 architectures is effective to enhance the SC performance significantly (the aud-vis baseline with 90.4\%), which is competitive with the fusion of high-complexity network architectures with using only visual data (90.5\%).
When we conduct PROD fusion of both audio and visual based frameworks (aud-vis PROD), we can achieve the best accuracy of 95.7\%.

\subsection{An application demo proposed}
\label{demo}

As we can achieve good results from high-performed deep learning frameworks mentioned in Section~\ref{all_fac}, we then conduct an application demo which uses a HTML front-end interface for uploading an input video and showing classification results on bar charts as shown in Figure~\ref{fig:F6}.
The back-end inference process of the demo is the aud-vis baseline with PROD fusion mentioned in Section~\ref{baseline}.
As input videos may have different lengths and scene context can be changed by time, bar charts present the classification accuracy on each 10-second segment instead of the entire recording time. 
By using the Docker software~\cite{docker}, the application of crowded scene classification is packaged to create a docker image for sharing\footnote{https://github.com/phamdanglam1986/An-application-demo-of-audio-visual-crowded-scene-classification-}.
Given the docker image, the application can be run on a wide range of computers with available docker software setup and easily integrated into any cloud-based system.
%The demo application also meets the real-time requirement with either cpu or gpu core.
 
\section{Conclusion}
\label{conclusion}
We have proposed an audio-visual dataset of five crowded scenes, then explored different benchmark frameworks on this dataset.
Our deep learning framework, which makes use of multiple spectrogram approach for audio input and fine-tuning strategy for visual input, achieves the best performance of 95.7\%. 
The results obtained from our experiments in this paper are very potential to further develop a complex system for detecting relevant-riot context.
Our future work is generating an audio-visual-text dataset which comprises both crowded scenes and daily scenes. Given the dataset, we can conduct comprehensive experiments, then propose a powerful indicator to detect a relevant-riot context.


\section*{Acknowledgement}
The AMMONIS project is funded by the FORTE program of the Austrian Research Promotion Agency (FFG) and the  Federal Ministry of Agriculture, Regions and Tourism (BMLRT) under grant no. 879705.
%\begin{thebibliography}{99}
\bibliographystyle{IEEEbib}
\bibliography{refs}
%\end{thebibliography}
\end{document}